# Deep learning approach for breast cancer diagnosis

Essam Rashed
Faculty of Informatics and Computer Science,
The British University in Egypt, Cairo, Egypt &
Faculty of Science, Suez Canal University, Ismailia, Egypt
essam.rashed@bue.edu.eg

M. Samir Abou El Seoud
Faculty of Informatics and Computer Science,
The British University in Egypt, Cairo, Egypt
Samir.elseoud@bue.edu.eg

## ABSTRACT
Breast cancer is one of the leading fatal disease worldwide with high risk control if early discovered. Conventional method for breast screening is x-ray mammography, which is known to be challenging for early detection of cancer lesions. The dense breast structure produced due to the compression process during imaging lead to difficulties to recognize small size abnormalities. Also, inter- and intra-variations of breast tissues lead to significant difficulties to achieve high diagnosis accuracy using hand-crafted features. Deep learning is an emerging machine learning technology that requires a relatively high computation power. Yet, it proved to be very effective in several difficult tasks that requires decision making at the level of human intelligence. In this paper, we develop a new network architecture inspired by the U-net structure that can be used for effective and early detection of breast cancer. Results indicate a high rate of sensitivity and specificity that indicate potential usefulness of the proposed approach in clinical use.

## 1. INTRODUCTION
Breast cancer is a major death leading worldwide with different occurrence reasons that is difficult to avoid in modern life especially in developed countries [1]. One of the major and common screening technologies is mammography which is widely available due to simple screening protocol and economic investigation cost. However, accurate diagnosis using mammograms requires a skillful expert radiologist that can identify abnormalities. In some cases, it is difficult to observe small size abnormal structures such as microcalcifications with size less than 1 mm. Moreover, dense breast structures might be difficult to distinguished from lesions with almost same structure and contrast. Therefore, there was always a need to develop a smart tool that provide expert-like analysis of mammograms as the early detection of small size lesion is likely increase the survival rate of the patient.

Recently, two new technology trends are leading a revolution in the development of better healthcare for the community. First, is the big data, where a huge amount of data can be easily available to be shared, easy to be processed and integrable to cover wide range of variabilities. A potential example in the field of medical imaging is the Cancer Imaging Archive [2], where a thousands of images with different modalities are available with almost full access. Second, is the deep learning, where artificial intelligence moved to a level close to the human intelligence (or beyond) [3]. Due to the availability of high computational power (mainly based on GPUs), it becomes possible to construct a deep neural networks with high number of layers that can extract huge number of features that was not possible before. Convolutional neural networks in particular leads to a remarkable impact in image analysis and understanding especially in image segmentation, classification and analysis [4]. Several models employ deep learning are already developed for diagnosis and identification of breast cancer through analysis of digital mammography [5-17].

In this paper, we propose a novel convolutional neural network (CNN) architecture for breast cancer diagnosis. The proposed architecture extract image features from different tracks of convolution/deconvolution layers. The proposed structure is designed to handle the problem of abnormality variation in digital mammography and it is expected to improve the detectability of lesion with higher quality compared to conventional methods

## 2. RELATED WORK
Development of automatic diagnosis system for breast cancer using mammography is a long time active track with achievements that employ different techniques to improve quality of diagnosis. Here, we briefly list recent trends that is based on the use of deep learning and convolutional neural networks. An interesting review is presented in [5], where several potential future tracks are discussed. For example, it discusses the problem of data availability and robustness of developed approaches to data acquisition methods. One potential promising approach is the deep learning as shown in [4]. In [6], a deep CNN is proposed using transfer learning. A set of Region-of-Interests are extracted from the mammograms and normalized before feed to the network. Results indicate high accuracy results that emphasis the applicability of transfer learning in this application. A cascade deep learning method is introduced in [7]. In this method, the problem of diagnosis is simplified into three stages as detection, segmentation and classification of mass. Although this approach is of relatively high computation cost compared to similar ones, but results indicate a notable detection quality improvement. A comparison study using relatively large amount of patient data is conducted in [8]. Several networks are evaluated and results indicate that GoogLeNet architecture can achieve highest accuracy value from area under the curve (AUC) measure measured value. A valuable measure for breast cancer risk analysis is the breast texture identification. However, another study reported that VGGNet results out performs GoogleNet and AlexNet [9]. These results indicate that it is important to have a uniform standard for evaluation include the image data and experiment parameters.

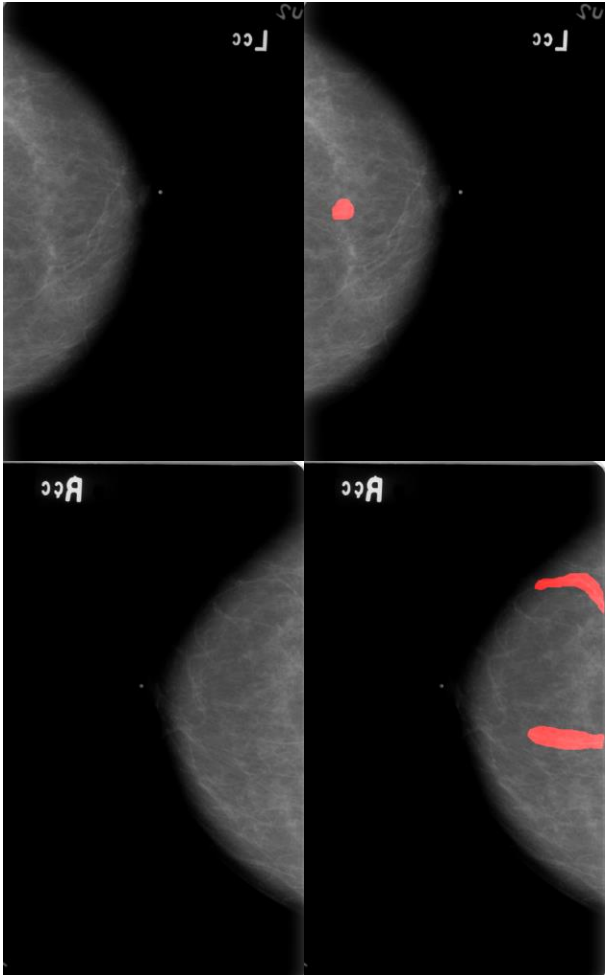

**Figure 1. Sample of mammograms from the CIBS-DDSM dataset Patient ID P_00038.**

In [10], deep learning is used for breast density segmentation and scoring of mammography texture. This task is implemented using unsupervised learning for texture features extraction. A two-step method is proposed for mammography diagnosis in [11]. A pre-processing is implemented for enhancement of image details followed by a deep CNN. The developed network architecture consists of CNN with fully connected network at the last layers. U-net architecture is used in the work presented in [12]. Results indicate a usefulness of U-net in mammogram diagnosis, however, several false positive results are reported. Interesting results are reported using asymmetric encoder-decoder network architecture [13]. From the research progress discussed above, we introduce a new network architecture that is inspired by U-net and modified to improve diagnosis results as detailed in next section.

## 3. MATERIALS AND METHODS

Automatic detection of breast abnormalities from mammograms is a challenging task due to the variation of size, texture, contrast of breast abnormalities. In this section, the dataset used in this study along with the network architecture is discussed.

### 3.1 Mammography Dataset

In this study, the Curated Breast Imaging Subset of Digital Database of Screening Mammography (CBIS-DDSM) [18] is used for both training and testing of the developed deep learning approach. This dataset contains mammography of 6,671 subjects of normal, benign, and malignant cases that is confirmed with pathology investigation. Images are provided in DICOM formats with binary labels for lesion identification. More details regarding the CBIS-DDSM dataset is available in [19]. The CBIS-DDSM dataset contains data for microcalcification and mass abnormalities of 603 (152) and 692 (202) subjects for training (testing), respectively. For each subject, images obtained from left and right breasts with bilateral craniocaudal (CC) and mediolateral oblique (MLO) views are available. We consider the use of CC views only and consider the combination of MLO views as future work. A sample mammograms used in this study is shown in Figure 1. As a pre-processing step, a set of Region-of-Interest (ROI's) represent the region with potential abnormality is selected from the original mammograms. The ROIs are selected to have the abnormality centered (as much as possible) within a 1024×1024 pixels. Histogram equalization process is used to normalize image contrast.

**Table 1. Network architecture of O-net.**

| Layer | Feature size | Layer | Feature size |
|---|---|---|---|
| input | 1024×1024 | | |
| 1 | 8×512×512 | 13 | 32×128×128 |
| 2 | 16×256×256 | 14 | 2×32×128×128 |
| 3 | 32×128×128 | 15 | 32×128×128 |
| 4 | 64×64×64 | 16 | 16×256×256 |
| 5 | 128×32×32 | 17 | 2×16×256×256 |
| 6 | 256×16×16 | 18 | 16×256×256 |
| 7 | 128×32×32 | 19 | 8×512×512 |
| 8 | 2×128×32×32 | 20 | 2×8×512×512 |
| 9 | 128×32×32 | 21 | 4×8×512×512 |
| 10 | 64×64×64 | 22 | 8×512×512 |
| 11 | 2×64×64×64 | 23 | 4×1024×1024 |
| 12 | 64×64×64 | output | 1024×1024 |

### 3.2 Network Architecture

The U-net is well-known CNN that is developed for image segmentation applications [20]. It is a composition of successive convolution and deconvolution layers as encoder-decoder architect. U-net provide an interesting architecture for image features extraction and used later in several imaging applications [4]. Here, we proposed the use of O-net, which is a combination of two U-nets connected at the encoding level and disconnected at the decoding. This architecture enables us to apply convolution/deconvolution processes with different kernel structures that aims at finding features with different size/contrast. The proposed architecture of O-net is shown in figure. 2 and details of feature size at each network layer is shown in Table 1. The convolution process is implemented with kernel size of (3×3) pixels, (1×1) padding and (1×1) stride, except for the layers from 9 to 19 where two branches (up and bottom) are exists. The up branch from layers 9 to 19 uses a larger convolution kernel with size (5×5) pixels, (2×2) padding and (1×1) stride. Pooling operations are set to maximum pooling of (2×2) pixels. Network input is 1024×1024 mammography and output is binary labels for abnormalities with the same image size. A set of data augmentation is used to generate additional pseudo training set. We consider a rotation of the original mammograms with random value with range 0-180° before ROI cropping. If rotation lead to include region outside the original domain of the original image, extended region is filled with zeros. If image markers (used to identify patient, breast or imaging position) is found in the cropped ROI, it is removed by zero filling.

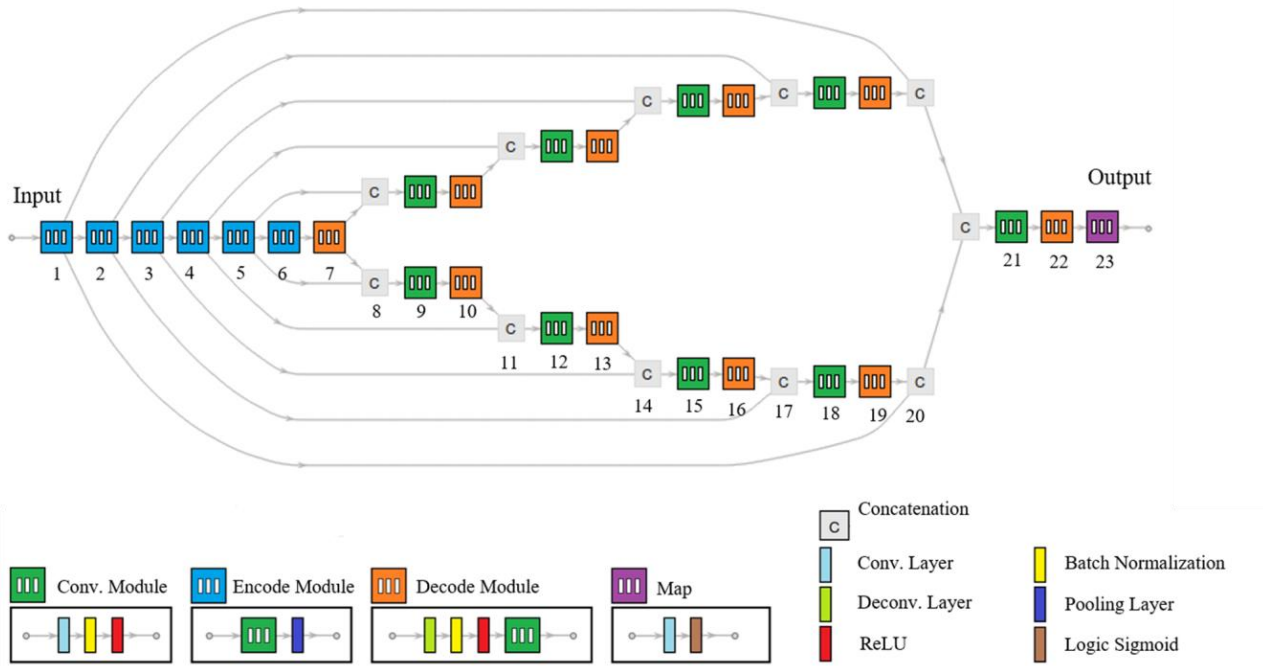

**Figure 2. The architecture of O-net with layer description below and layers are numbered.**

## 4. RESULTS

The proposed CNN is implemented using Wolfram Mathematica (R) 11.3 [21] on a workstation of 4×Intel (R) Xeon CPU @ 3.60 GHz, 64 GB memory and 3× *n*Vidia GeForce GTX 1080 GPU. Computations are conducted using GPUs to speed up the training. The network is trained for maximum 50 iterations with batch size of 4 images. The training can stop before reaching the maximum training iterations if the loss value difference is less than $10^{-3}$. We have used the default training optimizer algorithm, which is ADAM optimizer [22] and images are randomly shuffled before training. Two networks are trained, one is for the set of microcalcifications and the other is for the set of masses. The network training time is less than 100 minutes (pre-processing is excluded). An example of achieved results is shown in figure 3.

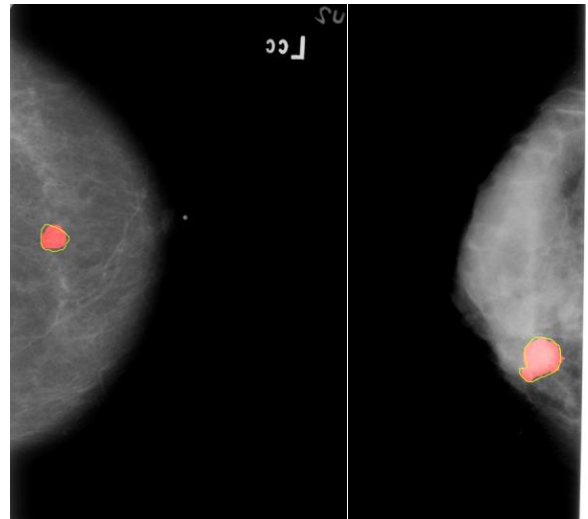

**Figure 3. Abnormality detection results for microcalcification (left) and mass (right). Red labeled region is golden truth label and region identified with yellow line is the network labeled region.**

The classification accuracy for microcalcification and masses are 94.31% and 95.01%. These results outperform reported results using well-known structures such as AlexNet, VGGNet and GoogleNet as reported in [9]. One potential reason for these results is the due to the combination of two different convolution kernels at the encoder tracks.

## 5. DISCUSSION

The proposed O-net structure shows a promising results in challenging diagnosis problem. However, these are just initial results that need further justification to clearly identify the performance optimized parameters. For example, a more complex network structure with additional kernel size may contribute to the improvement of classification accuracy. Also, the identification of training set is still computed manually by selecting specific ROIs that include the abnormalities. Automatic identification of the ROI, or processing the raw mammograms is more useful for practical and clinical use.

## 6. CONCLUSION

In this work, we presented a deep learning approach for the detection of breast cancer using mammograms. The proposed approach is developed following the development of convolution neural networks and it demonstrate how robust is deep learning in this application. There are several potential variations of the proposed network architecture that can be investigated and validated as a future work. It is still unclear what is the best network design that fit with the texture of digital mammograms and how sensitive is it. The proposed method can let to better performance of clinical use of breast cancer detection especially in early stages.